\documentclass{article}

\usepackage[T1]{fontenc}
\usepackage{microtype}
\usepackage{graphicx}
\graphicspath{ {./images/} }
\usepackage{booktabs}
\usepackage{multirow}
\usepackage[table]{xcolor}
\usepackage{colortbl}
\usepackage{amsfonts}
\usepackage{nicefrac}
\usepackage{listings}
\usepackage[linesnumbered,ruled,vlined]{algorithm2e}
\usepackage{enumitem}
\usepackage{csquotes}
\usepackage{lipsum}
\usepackage{arxiv}

\usepackage{hyperref}

\lstdefinestyle{mypython}{
  language=Python,
  backgroundcolor=\color{gray!10},
  basicstyle=\ttfamily\small,
  keywordstyle=\color{blue}\bfseries,
  commentstyle=\color{green!50!black}\itshape,
  stringstyle=\color{red},
  showstringspaces=false,
  frame=single,
  numbers=left,
  numberstyle=\tiny\color{gray},
  stepnumber=1,
  breaklines=true,
}

\definecolor{lightgreen}{rgb}{0.88, 1, 0.88}
\definecolor{lightgray}{gray}{0.95}
\definecolor{blue1}{RGB}{222,235,247}
\definecolor{blue2}{RGB}{198,219,239}
\definecolor{blue3}{RGB}{158,202,225}
\definecolor{blue4}{RGB}{107,174,214}
\definecolor{red1}{RGB}{254,224,210}
\definecolor{red2}{RGB}{252,187,161}
\definecolor{red3}{RGB}{252,146,114}
\definecolor{red4}{RGB}{222,45,38}

\title{CARMA: Comprehensive Automatically-annotated Reddit Mental Health Dataset for Arabic}
\author{
Saad Mankarious \quad Ayah Zirikly \\
School of Engineering and Applied Science \\
George Washington University, Washington, D.C. 20037 \\
\texttt{\{saadm, ayah.zirikly\}@gwu.edu}
}
\begin{document}

\maketitle

\begin{center}
\vspace{-0.5em}
\textbf{GitHub:} \href{https://github.com/fibonacci-2/carma}{\texttt{github.com/fibonacci-2/CARMA}} \quad
\raisebox{-0.2em}{\includegraphics[height=1em]{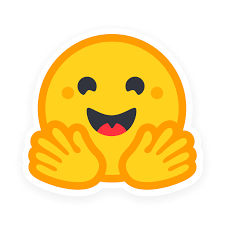}}\ 
\href{https://huggingface.co/datasets/smankarious/carma}{\texttt{huggingface.co/datasets/smankarious/carma}}
\vspace{1em}
\end{center}

\begin{abstract}
Mental health disorders affect millions worldwide, yet early detection remains a major challenge, particularly for Arabic-speaking populations where resources are limited and mental health discourse is often discouraged due to cultural stigma. While substantial research has focused on English-language mental health detection, Arabic remains significantly underexplored, partly due to the scarcity of annotated datasets. We present \textbf{CARMA}, the first automatically annotated large-scale dataset of Arabic Reddit posts. The dataset encompasses six mental health conditions, such as Anxiety, Autism, and Depression, and a control group. \textbf{CARMA} surpasses existing resources in both scale and diversity. We conduct qualitative and quantitative analyses of lexical and semantic differences between users, providing insights into the linguistic markers of specific mental health conditions. To demonstrate the dataset's potential for further mental health analysis, we perform classification experiments using a range of models, from shallow classifiers to large language models. Our results highlight the promise of advancing mental health detection in underrepresented languages such as Arabic.
\end{abstract}


\section{Introduction}
\begin{table}[htbp]
\centering
\small 
\begin{tabular}{@{}p{2.2cm}p{3.2cm}ccp{1.8cm}@{}}
\toprule
\textbf{Dataset} & \textbf{Target/Conditions} & \textbf{\# Posts} & \textbf{Source} & \textbf{Annotation} \\
\midrule
MentalQA \cite{alhuzali2025arahealthqa} & N/A & 157K & Altbbi Forum & Human \\
AraDepSu \cite{hassib-etal-2022-aradepsu} & Depression and suicide & 20K & Twitter & Human \\
CairoDep \cite{9694178} & Depression & 7K & Twitter & Human \\
Mood Change \cite{maghraby2022modern} & Depression & 49K & Twitter & Automatic \\
\midrule
CARMA (ours) & \begin{tabular}{@{}c@{}}ADHD, Anxiety, Autism, \\ Depression, OCD, Suicide\end{tabular} & 340K & Reddit & Automatic \\
\bottomrule
\end{tabular}
\vspace{5pt}
\caption{Comparison of the size of our dataset with prior Arabic mental health corpora. Existing resources exhibit limited breadth, primarily focusing on depression and suicide—and smaller scale, thereby constraining comprehensive mental health analysis in Arabic.}
\label{tab:data-size-comparison} 
\end{table}

\begin{figure}
    \centering
    \includegraphics[width=.8\textwidth]{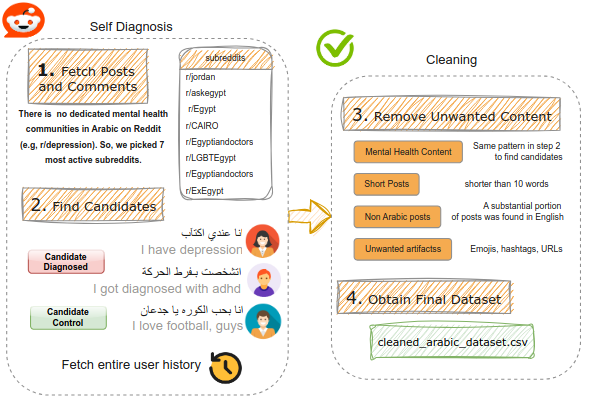}
    \caption{Key steps in dataset construction, beginning with raw Reddit data and culminating in the final datasets.}
    \label{fig:pipeline}
\end{figure}

Mental health conditions have a profound global impact on both society and the economy~\cite{santomauro2021global}. The World Health Organization (WHO) estimates that approximately 970 million people are affected worldwide, with up to 45\% of cases potentially going undiagnosed~\cite{downey2012undiagnosed}. These challenges are especially pronounced in Arabic-speaking populations, where cultural stigma and limited access to specialized care further exacerbate the issue~\cite{khatib2023understanding}.

Traditionally, mental health diagnoses rely on clinical interviews, self-report surveys, and standardized assessments. These approaches may fall short in capturing real-time or context-specific fluctuations in an individual's mental state. As a result, researchers are increasingly exploring alternative data sources, such as social media, to provide complementary, scalable, and more dynamic methods of mental health assessment. User-generated content on social media offers a rich, real-time stream of data that reflects users’ emotions, thoughts, and behaviors~\cite{reece2017instagram, park2013}. This has sparked growing interest in leveraging linguistic analysis of social media posts to identify patterns associated with mental health conditions~\cite{reece2017instagram}. Such approaches enable continuous mental health monitoring and early detection, providing means to reach individuals who may not otherwise engage with traditional mental health services~\cite{park2013}.

Prior work in mental health detection from social media has been predominantly focused on English-language data. Several large-scale datasets, such as SMHD~\cite{cohan2018smhd} and CLPsych~\cite{shing-etal-2018-expert, zirikly2019clpsych}, have enabled the development of robust models that achieve strong performance in identifying mental health conditions. However, equivalent resources for Arabic are scarce, with very limited work exploring Arabic-language user-generated content in this context~\cite{khatib2023understanding}. Furthermore, the existing datasets, most of which are curated from Twitter, focus exclusively on depression and suicide, neglecting other prevalent conditions such as ADHD and anxiety \cite{maghraby2022modern, hassib-etal-2022-aradepsu, 9694178, alhuzali2024mentalqa}. In addition, many of these datasets were collected some time ago, raising concerns that mental health trends on social media may have shifted, for example, due to the COVID-19 pandemic. This presents a significant research gap considering the linguistic diversity and unique cultural factors that influence mental health discourse in Arabic-speaking populations.

To address this gap, we present the first comprehensive, large-scale dataset based on Reddit that spans six mental health conditions. Our dataset significantly surpasses the scale and breadth of existing datasets in the literature, as Table \ref{tab:data-size-comparison} shows.

\paragraph{We summarize our contributions as follows:}
\begin{itemize}
\item \textbf{Introduction of CARMA:} We present \textbf{CARMA}, the first large-scale, automatically annotated Arabic dataset for mental health research, encompassing six mental health conditions and a control group, totaling over 340K posts.
\item \textbf{Condition-specific linguistic insights:} Through qualitative and quantitative analyses, we uncover distinctive linguistic and cultural markers associated with each mental health condition in Arabic discourse.
\item \textbf{Benchmarking and evaluation:} We establish strong baselines using both traditional classifiers and transformer-based models, demonstrating CARMA’s value as a benchmark resource for advancing Arabic mental health detection.
\end{itemize}

\section{Dataset Construction}
\label{sec-data-construction}

Our goal is to develop a large-scale Arabic dataset encompassing multiple mental health conditions to bridge the significant research gap in Arabic-language mental health detection. To this end, we propose two central hypotheses:

\textbf{Arabic Reddit communities contain abundant user data spanning multiple mental health conditions}
While Reddit lacks dedicated Arabic-language mental health forums comparable to English ones (e.g., r/ADHD or r/Depression), general Arabic communities such as r/Cairo and r/Arab may still host discussions reflecting users’ mental health experiences. This suggests that relevant linguistic signals can be extracted even from non-specialized Arabic Reddit spaces.

\textbf{Users with mental health conditions exhibit distinctive linguistic patterns}
Previous research demonstrates that linguistic behavior differs systematically between diagnosed individuals and control users~\cite{trifan2020understanding, cohan2018smhd, coppersmith2015adhd}. We extend this hypothesis to Arabic, conducting comprehensive linguistic and classification analyses to uncover interpretable signals that distinguish diagnosed users from controls.

To test these hypotheses, we follow the dataset construction pipeline illustrated in Figure~\ref{fig:pipeline}. We first collect diagnosed users through self-reported diagnosis patterns (Section~\ref{sec-diagnosed-dataset}), then curate a balanced control dataset (Section~\ref{control}). Finally, we apply rigorous cleaning and filtering steps, such as language and dialect, to ensure data integrity and robustness (Section~\ref{sec-cleaning}).

\begin{table*}[htbp]
\centering
\label{tab:desc-stats}
\begin{tabular}{lcccc}
\hline
\textbf{Condition} & \textbf{\# Users} & \textbf{\# Posts} & \textbf{Average Post Length} & \textbf{Std. Dev. Post Length} \\
\hline
Control & 4,086 & 270,622 & 44.87 & 64.93 \\
\hline
ADHD & 41 & 6,186 & 44.22 & 70.31 \\
Anxiety & 187 & 22,306 & 46.09 & 66.11 \\
Autism & 38 & 6,404 & 49.85 & 91.42 \\
Depression & 171 & 21,085 & 49.32 & 70.94 \\
OCD & 60 & 8,334 & 45.34 & 63.20 \\
Suicide & 36 & 4,613 & 51.07 & 82.24 \\
\hline
\end{tabular}
\caption{User and post statistics by mental health condition and control group in the final cleaned dataset.}
\end{table*}

\subsection{Diagnosed Dataset}
\label{sec-diagnosed-dataset}
We used the Arctic Shift API\footnote{\url{https://github.com/ArthurHeitmann/arctic_shift}} to download data from eight general subreddits\footnote{We targeted the most active communities on Reddit, ensuring it covers multiple dialects such as Egyptian and Levantine: \texttt{r/Egypt, r/jordan, r/askegypt, r/CAIRO, r/Egyptiandoctors, r/LGBTEgypt, r/SaudiForSaudis and r/ExEgypt}} that were chosen for their high Arabic posting activity, due to the lack of dedicated mental health Arabic subreddits (e.g., \texttt{r/ADHD} or \texttt{r/Bipolar} for English speaking users).

We then applied algorithm \ref{alg:proximity_matching} to identify potential diagnosed users using a curated diagnosis pattern keywords and diagnose phrases. This algorithm is inspired by self-reported diagnosis pattern proposed by \cite{coppersmith2015adhd} for identifying diagnosed users from an English corpus. We use the enhancement from \cite{cohan2018smhd}  of introducing a distance threshold $d$ between the mention of the diagnosis phrase and the appearance of the condition, which was found to be 40 characters or less. This is to allow flexibility and capture any occurrences we can get which is crucial due to scarcity of data. 

The algorithm uses a curated list of diagnosis keywords $K$ (e.g., "ana itshakaṣtu bi-"—I got diagnosed with) and a diagnosis phrase $P$ (e.g., "ikti'ab ḥadd"—Severe Depression) and check if they happen within the optimal distance (40 characters). To enhance the inclusivity and robustness of diagnosis lists, we incorporated variations across different Arabic dialects including Modern Standard Arabic, Egyptian and Levantine, along with frequent spellings and misspellings to account for the diverse nature of posting habits social media users. Check appendix \ref{appendix-self-diagnosis} for a snippet from those lists.

Applying the algorithm on the downloaded data resulted in finding a total of 959 potential diagnosed users from the targeted subreddits, as shown in Table \ref{tab:subreddit_counts}. In the next step, we retrieved all of those users' content on Reddit and passed it along the cleaning pipeline, as Figure \ref{fig:pipeline} shows. 

\textbf{Self-Diagnosis Pattern Validation} To evaluate labeling accuracy, we manually reviewed a random sample of 100 matched posts. Of these, 83 were valid self-diagnoses, while 17 were false positives, typically describing another person’s diagnosis (e.g., “my brother got diagnosed with ADHD”) or negated statements (“I was not diagnosed with ADHD”). Similar challenges in self-reported labeling have been observed in English datasets, including suicide-related corpora~\cite{zirikly2019clpsych, cohan2018smhd}.

\textbf{Language and Dialect Detection} We used a BERT-based model\footnote{\url{https://huggingface.co/lafifi-24/arabert_arabic_dialect_identification}} to identify the distribution of Arabic dialects in the dataset. As shown in Figure~\ref{fig:dialects}, Egyptian Arabic dominates due to the high participation of Egyptian users on Reddit, followed by Jordanian (JO), Saudi (SA), and Palestinian (PL) dialects. This distribution aligns with population trends, where Egypt’s larger population (approximately 116M) naturally contributes to greater online activity compared to Saudi Arabia (35M) and Jordan (11M).\footnote{Population estimates sourced from official 2024 UN demographic data.}

\begin{algorithm}[t]
\caption{Self-reported Diagnosis via Phrase–Keyword Matching}
\label{alg:proximity_matching}
\KwIn{Reddit Post $C$, Diagnosis Phrase $P$, Diagnosis Keyword $K$, Distance Threshold $d$}
\KwOut{Set of matched phrase–keyword pairs}

\BlankLine
\Begin{
  Find all indices of $P$ in $C$ $\rightarrow$ phrase\_matches\;
  Find all indices of $K$ in $C$ $\rightarrow$ keyword\_matches\;
  
  \ForEach{$(p_{s}, p_{e}) \in$ phrase\_matches}{
    \ForEach{$(k_{s}, k_{e}) \in$ keyword\_matches}{
      \If{$|p_{s}-k_{s}| \leq d$ \textbf{or} $|k_{e}-p_{e}| \leq d$}{
        Record $(P,K)$ as match\;
      }
    }
  }
  \KwRet all matches\;
}
\end{algorithm}

\subsection{Control Dataset}
\label{control}
To construct the control dataset, we collected around~\textbf{1M} posts from the same eight subreddits used for identifying diagnosed users, ensuring consistency in posting behavior across both groups. Any user who exhibited signs of mental health in any of their posts, detected using Algorithm \ref{alg:proximity_matching}, was immediately excluded. This step is critical to minimize the risk of including potentially affected individuals in the control group and to prevent the model from learning spurious correlations, which could lead to overly optimistic and misleading performance~\cite{cohan2018smhd, zirikly2019clpsych, shing-etal-2018-expert}. The remaining users’ posts were subsequently processed using the same cleaning pipeline used for diagnosed users content, described in Section~\ref{sec-cleaning}.

\subsection{Data Cleaning}
\label{sec-cleaning}
We used the \texttt{langdetect} library~\cite{ercdidip2022} to filter out non-Arabic content from the dataset, which was common due to the frequent use of English among native Arabic speakers online. This process excluded 30\% of the diagnosed dataset (approximately 140K posts out of the originally collected 480k) and 13\% of the control dataset. Despite this severe reduction in data size, we argue that this step was essential to preserve a coherent Arabic linguistic signal within the dataset. Additionally, we removed posts containing fewer than 10 words to ensure that each entry provided sufficient contextual information. Further cleaning involved removing mental health disclosure content, as detailed in Section~\ref{control}.

\begin{table*}[htbp]
\centering
\small
\begin{tabular}{lllll}
\toprule
\textbf{Condition} & \textbf{1-gram Terms} & \textbf{TF-IDF Diff} & \textbf{2-gram Terms} & \textbf{TF-IDF Diff} \\
\midrule

\multirow{5}{*}{ADHD} & "anta" (You) & \cellcolor{red3} -0.0067 & "'ashan fi" (Because in) & \cellcolor{red2} -0.0029 \\
 & "suriya" (Syria) & \cellcolor{red2} -0.0030 & "'ibara 'an" (Consists of) & \cellcolor{red2} -0.0031 \\
 & "an-nihaya" (The end) & \cellcolor{red2} -0.0017 & "'ala al-aqall" (At least) & \cellcolor{red1} -0.0041 \\
 & "as-salam" (Peace) & \cellcolor{red3} -0.0050 &  &  \\
\midrule

\multirow{5}{*}{Anxiety} & "'andi" (I have) & \cellcolor{blue3} 0.0175 & "bass ana" (But I) & \cellcolor{blue3} 0.0168 \\
 & "bass" (Just/But) & \cellcolor{blue4} 0.0415 & "ana 'andi" (I have) & \cellcolor{blue3} 0.0132 \\
 & "nafsi" (Myself) & \cellcolor{blue3} 0.0138 & "ana mish" (I am not) & \cellcolor{blue3} 0.0190 \\
 & "suriya" (Syria) & \cellcolor{red3} -0.0031 & "masr mish" (Egypt not) & \cellcolor{red1} -0.0024 \\
\midrule

\multirow{5}{*}{Autism} & "kadhalik" (Also) & \cellcolor{red2} -0.0018 & "akthar min" (More than) & \cellcolor{red4} -0.0096 \\
 & "as-salam" (Peace) & \cellcolor{red3} -0.0048 & "aw hatta" (Or even) & \cellcolor{red3} -0.0043 \\
 & "madi" (Financial) & \cellcolor{red1} -0.0014 & "hal fi" (Is there) & \cellcolor{red4} -0.0087 \\
 & "nahiya" (Aspect) & \cellcolor{red2} -0.0034 & "'ala an-nas" (On people) & \cellcolor{red3} -0.0049 \\
 & "jama'a" (Group) & \cellcolor{red2} -0.0036 & "'ala hasab" (Depending on) & \cellcolor{red4} -0.0102 \\
\midrule

\multirow{5}{*}{Depression} & "'andi" (I have) & \cellcolor{blue4} 0.0214 & "'andi ikti'ab" (I have depression) & \cellcolor{blue3} 0.0153 \\
 & "lan" (Will not) & \cellcolor{red3} -0.0020 & "ani mish" (I am not) & \cellcolor{blue3} 0.0131 \\
 & "ana" (I) & \cellcolor{blue4} 0.0599 & "bass ana" (But I) & \cellcolor{blue3} 0.0163 \\
 & "ikti'ab" (Depression) & \cellcolor{blue3} 0.0119 & "hatta fi" (Even in) & \cellcolor{red2} -0.0023 \\
\midrule

\multirow{5}{*}{OCD} & "laysa" (Not) & \cellcolor{red3} -0.0037 & "kull al-li" (All that) & \cellcolor{red4} -0.0078 \\
 & "as-safar" (Travel) & \cellcolor{red3} -0.0023 & "ma rah" (Will not) & \cellcolor{red3} -0.0057 \\
 & "al-jami'at" (Universities) & \cellcolor{red3} -0.0020 & "anta fi" (You are in) & \cellcolor{red2} -0.0037 \\
 & "al-jadid" (The new) & \cellcolor{red3} -0.0022 & "kabir jiddan" (Very big) & \cellcolor{red2} -0.0023 \\
\midrule

\multirow{5}{*}{Suicide} & "idha" (If) & \cellcolor{red4} -0.0273 & "bass ma" (But not) & \cellcolor{red4} -0.0173 \\
 & "wash" (What-dialect) & \cellcolor{red4} -0.0094 & "al yin" (The-incomplete) & \cellcolor{red4} -0.0059 \\
 & "adri" (I don't know) & \cellcolor{red3} -0.0029 & "hal anta" (Are you) & \cellcolor{red2} -0.0036 \\
 & "lak" (For you) & \cellcolor{red4} -0.0110 & "as-salam 'alaykum" (Peace be upon you) & \cellcolor{red4} -0.0145 \\
\bottomrule
\end{tabular}
\caption{Significant terms by condition against the control group with 1-gram and 2-gram terms. Blue colors indicate positive TF-IDF differences (more frequent in diagnosed group), red colors indicate negative differences (more frequent in control group). Color intensity corresponds to magnitude of difference. All values are statistically significant as per a t-test with ***$p\_adj < 0.001$, **$p\_adj < 0.01$, *$p\_adj < 0.05$.}
\end{table*}


\section{Previous Work}

\textbf{Arabic Mental Health Detection}
The task of detecting mental health status from social media data has been explored extensively in English. One notable example is the UMD dataset~\cite{shing-etal-2018-expert, zirikly2019clpsych}, which consists of manually annotated English Reddit data for assessing suicide risk. In that work, researchers demonstrated that combining lexicon-based features with features generated by large language models (LLMs) produced strong results across various machine learning models. Building on this foundation, recent studies have applied LLMs to Reddit-based mental health detection tasks, showing promising performance. This trend is reflected in recent shared tasks from CLPsych 2024 and 2025~\cite{chim-etal-2024-overview, tseriotou-etal-2025-overview}, which focus on leveraging LLMs to identify indicators of suicidality and other mental health conditions in online posts.

\textbf{Arabic Mental Health Datasets}
Several attempts have been made to build Arabic mental health datasets, primarily using Twitter data. AraDepSu \cite{hassib-etal-2022-aradepsu}, CairoDep \cite{9694178}, and \cite{maghraby2022modern} constructed datasets through both human annotation and automatic labeling. These efforts produced relatively large corpora that proved effective for specific tasks, such as answering the Patient Health Questionnaire-9 (PHQ-9) to detect depression. \cite{maghraby2022modern}, for example, used an algorithmic approach to automatically categorize Twitter data into nine groups, while CairoDep employed manual labeling of a depression-specific corpus. Furthermore, MentalQA curated a dataset based on the Saudi medical forum Altebbi, adopting a QA-style format to support mental health-related tasks.  In a recent shared task built on this resource, teams designed systems capable of answering user medical questions in Arabic. Team Sindbad, for instance, achieved top performance of 0.67 BERTScore using GPT 3.5-turbo \cite{a-morsy-etal-2025-sindbad, alhuzali2025arahealthqa}.

Despite growing interest in mental health detection, resources for Arabic remain scarce due to the lack of large, well-annotated datasets. Existing efforts primarily target depression, with a few addressing suicidality, leaving other prevalent conditions such as anxiety, ADHD, and autism largely overlooked. Moreover, most prior datasets are constrained in scale and diversity because they rely on manual labeling, which limits both coverage and reproducibility \cite{zirikly2019clpsych, maghraby2022modern, hassib-etal-2022-aradepsu, 9694178, alhuzali2024mentalqa}. To our knowledge, no comprehensive Arabic dataset currently exists that captures multiple mental health conditions at scale, including underrepresented ones such as anxiety, ADHD, and OCD. This gap critically impedes progress in developing and evaluating Arabic mental health detection models, highlighting the need for a large, reproducible benchmark dataset to enable inclusive and data-driven research in this domain.

\textbf{Self-Reported Diagnosis}
To construct a large-scale dataset for Arabic mental health analysis, we adopt automatic labeling through self-reported diagnosis. Originally introduced by \cite{coppersmith2015adhd} and refined in later studies~\cite{yates2017depression, cohan2018smhd}, this method automatically identifies users who disclose a mental health condition by matching their text against carefully curated diagnostic patterns and keywords. For example, \cite{cohan2018smhd} developed a large Reddit-based dataset of English posts covering nine mental health conditions using this fully automated approach. The dataset included a substantial control group and required no human annotation, thereby overcoming major scalability barriers and enabling reproducible mental health research.

Beyond English, self-reported diagnosis has been successfully adapted to other languages. \cite{zanwar2023smhd}, for instance, recreated the SMHD dataset for German by tailoring diagnosis patterns to local linguistic structures. Their work produced a dataset spanning six mental health conditions and a control group, achieving an F1 score of \textbf{56.12} for bipolar disorder classification using the PsyLing model. This cross-linguistic applicability demonstrates the robustness and flexibility of the self-reported diagnosis framework, motivating its extension to underrepresented languages such as Arabic.
\vspace{-20pt}
\section{Dataset Exploration}
\vspace{-10pt}
\label{sec-data-exploration}
In the following sections we explore in depth different quantitative and qualitative aspects of the dataset. We also motivate a binary classification experiment to demonstrate the potential of the dataset to detect mental illness for Arabic audience.

Table~\ref{tab:desc-stats} summarizes the key descriptive statistics of the final dataset after cleaning (see subsection~\ref{sec-cleaning}). The dataset includes 340,343 posts, spanning 4,629  control users and 543  users across six mental health conditions. The control group accounts for the majority of posts, providing a solid baseline for comparative analysis.

Among the diagnosed groups, anxiety and depression were the most represented conditions. Autism and ADHD also showed relatively high average post counts per user, suggesting higher user activity within those groups.

\vspace{-20pt}
\subsection{Descriptive Analysis}
\label{sec-des-analysis}
To examine the semantic characteristics of the collected data, we conducted a TF-IDF bag-of-words analysis using both unigrams and bigrams. In addition, we generated word clouds to visually compare linguistic patterns between the diagnosed and control groups, providing an initial qualitative overview of mental health discourse in Arabic Reddit communities.

\begin{figure}
\centering
\includegraphics[width=1\linewidth]{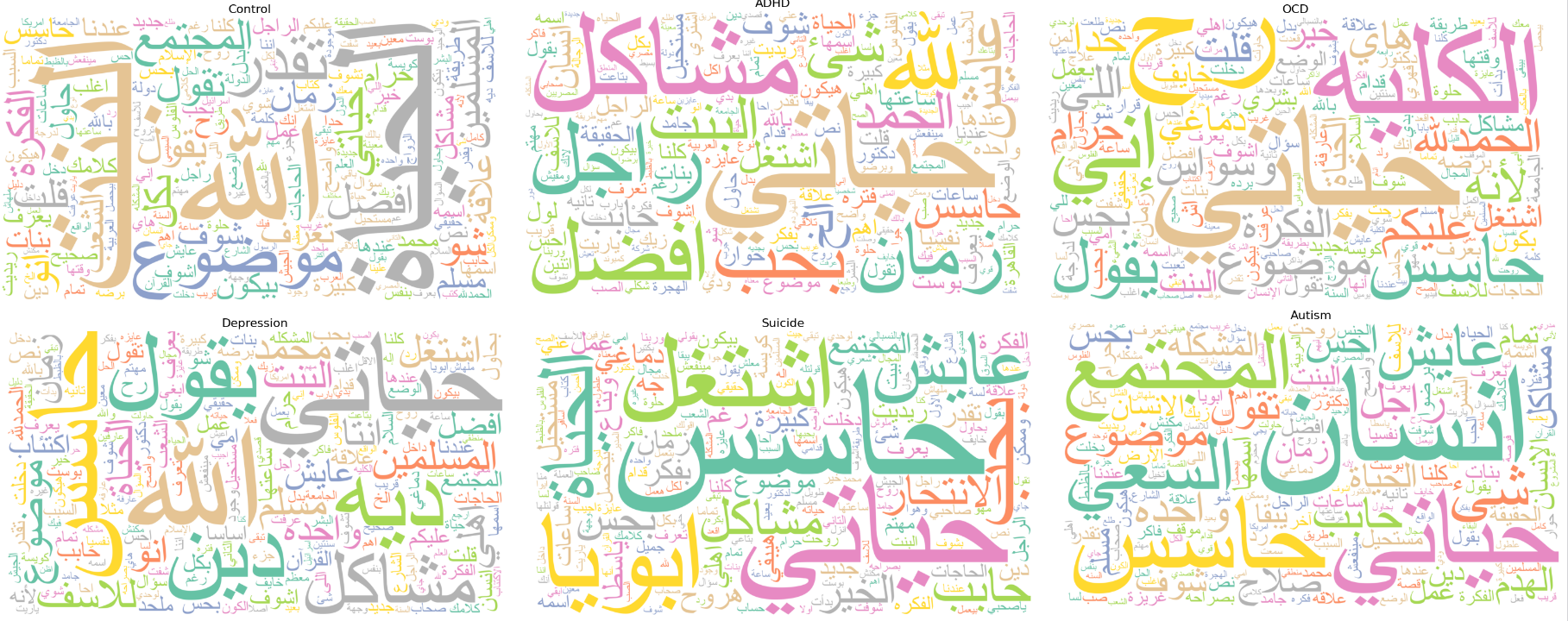}
\caption{Descriptive language of different groups in the dataset.}
\label{fig:wordclouds}
\end{figure}

As shown in Figure~\ref{fig:wordclouds}, all diagnosed classes use highly personal language directly related to mental health, such as "hayati"-my life, "hases"-I feel, and "nafsi"-myself. In contrast, such expressions are rare in the control group, where users more often employ impersonal or passive forms, e.g., "al-hayah"-the life.


Each diagnosed condition exhibits distinctive linguistic and semantic patterns that align with previously documented psychological and linguistic behaviors. Using both TF-IDF and word cloud analyses, we observed clear lexical and semantic distinctions across the five main classes compared to the control group.

\textbf{ADHD:}
Posts associated with ADHD frequently include interpersonal and evaluative terms such as "anta"-You, "'ala al-aqall"-At least, and "an-nihaya"-The end. These expressions, identified through TF-IDF analysis, suggest users often describe personal experiences and comparative thoughts, possibly reflecting cognitive patterns such as impulsivity and topic switching. Similar discourse behaviors, marked by disorganized focus and shifting narrative attention, were observed in English ADHD corpora~\cite{coppersmith2015adhd, cohan2018smhd}.

\textbf{Anxiety:}
Anxiety-related posts are dominated by first-person and introspective language, with TF-IDF revealing frequent use of "'andi"-I have, "bass ana"-But I, and "nafsi"-Myself. This pattern, supported by the word cloud analysis, indicates high self-focus and emotional rumination. Prior studies~\cite{trifan2020understanding, cohan2018smhd} reported similar tendencies, where anxiety-diagnosed users exhibited increased self-referential and negative-emotion words, consistent with heightened internal awareness and avoidance phrasing seen in terms like "ana mish"-I am not.

\textbf{Autism:}
Language in autism-related posts displays a detached and analytical tone. High TF-IDF terms such as "kadhalik"-Also, "akthar min"-More than, and "nahiya"-Aspect reveal a preference for structured and comparative phrasing. References to social contexts, such as "'ala an-nas"-On people and "jama'a"-Group, reflect discussions about interpersonal experiences rather than emotional sharing. This aligns with prior cross-condition findings~\cite{cohan2018smhd}, which showed that autism posts tend to exhibit lower emotional valence and higher cognitive process words.

\textbf{Depression:}
Depression-related posts show strong self-disclosure and explicit diagnostic phrasing, with TF-IDF highlighting "'andi ikti'ab"-I have depression, "ikti'ab"-Depression, and "ana"-I as frequent terms. This pattern reflects direct reference to one's condition, consistent with the self-reporting behavior noted in English corpora~\cite{cohan2018smhd, zirikly2019clpsych}. The recurrent negation forms, such as "ani mish"-I am not, further indicate expressions of self-doubt and emotional withdrawal typical of depressive discourse.

\textbf{Suicide:}
Posts indicative of suicidal ideation are characterized by conditional and existential phrasing, including "idha"-If, "adri"-I don't know, and "hal anta"-Are you, as seen in the TF-IDF output. These terms signal uncertainty and questioning, often associated with cognitive indecision and emotional distress.

\textbf{Control Group}
Using the same list of stop words, the control group displays general language reflecting commonly discussed topics on Reddit. This includes terms such as "mawdu'"-topic and "allah"-God, where users are typically seeking general advice or engaging in religious discussions. The absence of mental health terminology in the control group is encouraging for downstream mental health analysis, as models trained on this data can effectively differentiate between typical users and those exhibiting mental health concerns~\cite{cohan2018smhd}. This distinction results from the rigorous cleaning process, including the removal of both mental health–related and non-Arabic content from the training set.


\subsection{Binary Classification}
To evaluate our second hypothesis from Section~\ref{sec-data-construction}, we trained binary classifiers on the collected diagnosed and control data to predict each diagnosed label (\texttt{adhd}, \texttt{autism}, \texttt{anxiety}, \texttt{depression}, \texttt{ocd}, or \texttt{suicide}). This is a single-label balanced classification where we sample same number of posts from diagnosed and control datasets. We do not address multi-label classification due to lack of comorbidity in the dataset. 

\textbf{Transformer Embeddings} Dense vector representations of varying dimensionality (depending on model used) were generated using four language models. 
\begin{itemize}[nosep]
    \item Arabic-labse~\cite{nacar2024enhancingsemanticsimilarityunderstanding}
    \item AraBERT~\cite{abdul-mageed-etal-2021-arbert}
    \item MARBERT~\cite{abdul-mageed-etal-2021-arbert}
    \item CAMeLBERT~\cite{inoue-etal-2021-interplay}
\end{itemize}
These embeddings were then used to train five classifiers: Logistic Regression, MLP, SVM, AdaBoost, and Random Forest.\footnote{See Appendix \ref{training-params} for training configurations.} Detailed results are provided in Table \ref{tab:binary-classification-1}.

\textbf{Finetuning} Our second classification task was achieved by fine-tuning six Arabic-optimized transformer models end-to-end, updating both the pre-trained linguistic representations and the task-specific classification layers using mental health annotations (see Appendix \ref{training-params} for the full list). This approach enables the models to capture condition-specific textual patterns while retaining general Arabic language understanding capabilities~\cite{araujo-etal-2021-multilingual}. Results of this expirment are reported in Table \ref{tab:direct-finetuning}

These classification signals can be leveraged for similar diagnostic tasks or analyzed independently to uncover language patterns associated with each mental health condition. We conducted binary classification at both the \textit{post level} and the \textit{user level}, ensuring a balanced 1:1 ratio between each diagnosed group and its corresponding control group.

\section{Results and Discussions}

\begin{table*}[htbp]
\centering
\label{tab:binary-classification-1}
\begin{tabular}{llllcccc}
\toprule
\textbf{Condition} & \textbf{Classification Model} & \textbf{Embedding Model} & \textbf{Accuracy} & \textbf{Precision} & \textbf{Recall} & \textbf{F1-Score} \\
\midrule
ADHD & Decision Tree & Arabic-labse & 0.76 & 0.70 & \textbf{0.88} & 0.78 \\
Anxiety & SVM & Arabic-labse & 0.83 & 0.82 & 0.84 & \textbf{0.83 }\\
Autism & Gradient Boosting & CAMeLBERT & 0.81 & 0.78 & \textbf{0.88} & \textbf{0.82} \\
Depression & Gradient Boosting & AraBERT & 0.71 & 0.71 & 0.71 & 0.71 \\
OCD & Gradient Boosting & CAMeLBERT & 0.83 &\textbf{ 0.90} & 0.75 & \textbf{0.82} \\
Suicide & Logistic Regression & AraBERT & 0.73 & 0.80 & 0.57 & 0.67 \\
\bottomrule
\end{tabular}
\caption{Binary Classification Results for Various Mental Health Conditions using Different Models trained on Transformer Embeddings. This is a 1 to 1 user level classification between the control group and each diagnosed group. Control users are sampled randomly with replacement.}
\end{table*}

\begin{table*}[htbp]
\centering
\label{tab:direct-finetuning}
\begin{tabular}{@{}l>{\raggedright\arraybackslash}p{8cm}ccc@{}}
\toprule
\textbf{Condition} & \textbf{Model} & \textbf{F1 Score} & \textbf{Precision} & \textbf{Recall} \\
\midrule
Suicide & \texttt{aubmindlab/bert-base-arabertv02-twitter} & \textbf{0.37} & 0.29 & \textbf{0.52} \\
ADHD & \texttt{CAMeL-Lab/bert-base-arabic-camelbert-mix} & 0.30 & \textbf{0.31} & 0.28 \\
Autism & \texttt{CAMeL-Lab/bert-base-arabic-camelbert-mix} & \textbf{0.33} & \textbf{0.29} & 0.40 \\
OCD & \texttt{CAMeL-Lab/bert-base-arabic-camelbert-mix} & 0.29 & 0.27 & 0.31 \\
Depression & \texttt{facebook/muppet-roberta-large} & 0.21 & 0.13 & \textbf{0.54} \\
Anxiety & \texttt{UBC-NLP/MARBERTv2} & 0.21 & 0.22 & 0.21 \\
\bottomrule
\end{tabular}
\caption{Post Level Binary Classification through Fine-Tuning Pretrained Models Directly on Arabic Data.}
\end{table*}

\textbf{Self-Reported Diagnosis Enabled a Comprehensive Arabic Mental Health Dataset} We demonstrated that self-reported diagnosis can be effectively applied to Arabic social media to identify multiple mental health conditions, aligning with prior English-language efforts \cite{coppersmith2015adhd, cohan2018smhd, zanwar2023smhd}. CARMA surpasses previous datasets in both scale and diversity, encompassing ten mental health conditions, compared to SMHD’s eight \cite{cohan2018smhd}. To ensure data reliability, we implemented language and dialect filtering, significantly improving corpus integrity.

\textbf{Neglected Mental Health Conditions Are Linguistically Distinct in Arabic Communities} CARMA reveals that conditions often underexplored in Arabic contexts, such as Anxiety, OCD, and ADHD, are highly represented and linguistically distinguishable. Our lexical analysis identified clear divergences between diagnosed and control groups, paralleling findings in English corpora \cite{yates2017depression, cohan2018smhd} These results suggest that Arabic speakers express mental distress through culturally grounded terms, often interwoven with religious and gendered discourse.

\textbf{Distinct Linguistic Patterns Differentiate Mental Health Conditions} Across conditions, we observed distinct linguistic markers that reflect both shared emotional expression and culture-specific framing. For example, Anxiety-related posts often contained first-person pronouns and affective terms, while OCD discourse exhibited heightened self-referential and moral language, reflecting cultural attitudes toward responsibility and faith.

\textbf{CARMA Enables Accurate and Scalable Detection of Mental Health Conditions} Model evaluation (Table~\ref{tab:direct-finetuning}) shows that \texttt{SVM} classifier achieved the highest F1 score (0.83) for Anxiety detection on embeddings from \texttt{Arabic-labse} embeddings. These results highlight the effectiveness of Arabic-specific transformers in low-resource mental health detection and affirm the potential of social media data for scalable, early identification of mental health signals in underrepresented languages. 

\section{Conclusion}
In this work, we introduced the first large-scale automatically annotated dataset for Arabic mental health research, constructed from Reddit posts and spanning six mental health conditions alongside a control group. By leveraging self-reported diagnosis patterns previously established in English-language studies, we provided a reliable and rigorously cleaned resource for investigating the linguistic markers of mental health in Arabic. Our analyses demonstrated that the dataset captures meaningful lexical and semantic distinctions between diagnosed and control users, and our experiments showed its utility across both hybrid classification pipelines and fine-tuned transformer models. The results underscore the dataset’s scalability and effectiveness for building predictive tools, while also highlighting the presence of underexplored conditions such as OCD and ADHD in online Arabic communities. Beyond immediate model performance, this dataset lays the groundwork for advancing inclusive and representative NLP for mental health, offering researchers a valuable resource for early detection, intervention, and the broader study of mental health discourse in underrepresented languages. In future work, we aim to further build on this work by analyzing comorbidity between conditions in light of new data collected. Furthermore, we plan to incorporate mental health specific lexical analysis tools such as LIWC should they become available.

\section{Limitations}

This study has several limitations that warrant consideration. First, the dataset primarily reflects social media users from a limited demographic, which may be biased towards male participants and certain dialects, thereby limiting the generalizability of our results. Second, although our models show encouraging performance, they do not fully capture the complexity and subtleties of mental health expressions in natural language, including contextual and cultural factors.

\section{Ethical Considerations}

Utilizing social media data for mental health research entails important ethical concerns, particularly around informed consent and user privacy. Although Reddit posts are publicly accessible\footnote{Reddit permits third-party access to public content via APIs (see \url{https://www.reddit.com/policies/privacy-policy} for details).}, users may not have explicitly consented to their data being used for research. We therefore adopted rigorous measures to anonymize data and protect user confidentiality. Our research adheres to ethical guidelines by allowing data access through Data Usage Agreement, similarly to what has been established in previous datasets like SHMD, avoiding any attempts to identify individuals, and emphasizing responsible use of findings to benefit mental health understanding without compromising privacy.

\bibliographystyle{unsrt}  
\bibliography{references}

\appendix
\section{Appendix}
\label{sec:appendix}

\begin{table}
\centering
\begin{tabular}{l c c}
\hline
Artifact & Total Count & Affected (\%) \\
\hline
Emojis        & 38,641  & 28,952 (10.6) \\
Numbers       & 100,515 & 40,861 (15.0) \\
English Words & 197,716 & 36,009 (13.2) \\
URLs          & 8,080   & 5,873 (2.2)  \\
Hashtags      & 4,355   & 718  (0.3)  \\
Mentions      & 289     & 170 (0.1)  \\
\hline
\end{tabular}
\vspace{5pt}
\caption{Unwanted artifacts detected in the dataset and percentage of affected posts. All such artifacts were removed from the dataset.}
\label{tab:artifact-summary}
\end{table}
\subsection{Data Statistics}
\begin{figure}
    \centering
    \includegraphics[width=.5\linewidth]{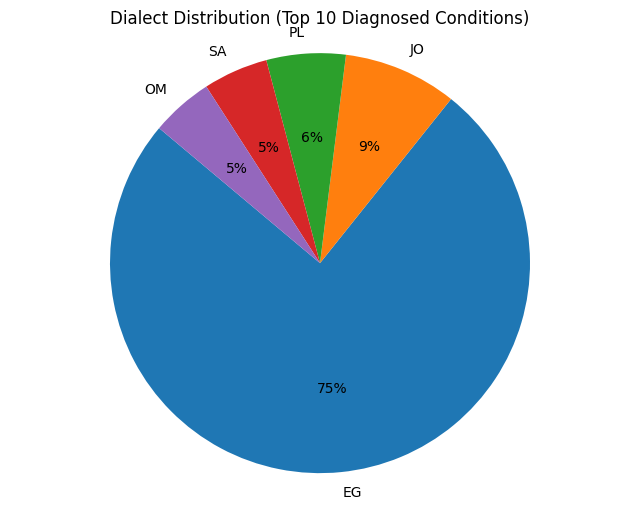}
    \caption{Dialects represented in the dataset.}
    \label{fig:dialects}
\end{figure}
\begin{table}[htbp]
\centering
\caption{Counts of diagnosed posts identified by the algorithm from each subreddit.}
\label{tab:subreddit_counts}
\begin{tabular}{lc}
\toprule
\textbf{Subreddit} & \textbf{Post Count} \\
\midrule
Egyptiandoctors & 20 \\
askegypt & 23 \\
arabs & 2 \\
LGBTEgypt & 11 \\
CAIRO & 280 \\
jordan & 110 \\
Egypt & 143 \\
ExEgypt & 237 \\
SaudiForSaudis & 133 \\
\bottomrule
\end{tabular}
\end{table}

\subsection{Diagnosis Patterns}
\label{appendix-self-diagnosis}
Here are the lists of condition specific keywords and diagnosis phrases used identify diagnosed users. Same patterns were used to clean the dataset from mental health content.
\textbf{Standard Arabic Phrases}
"ana musab bi-" – I am diagnosed with, "u'ani min" – I suffer from, "tashkhisi huwa" – My diagnosis is, "tumma tashkhisi bi-" – I was diagnosed with,
"tashkhis at-tabib huwa" – The doctor's diagnosis is, "tumma at-ta'kid 'ala isabati bi-" – It was confirmed that I have

\textbf{Egyptian Dialect Variations}
"ana 'andi" – I have, "'andi" – I have, "ana kunt 'ind ad-duktur wa-shakhhasani bi-" – I went to the doctor and was diagnosed with, "ad-duktur qalli 'andak" – The doctor told me you have

\textbf{Variants with Dialectical Pronunciations}
"ana shakhasuni bi-" – They diagnosed me with, "shakhasani ad-duktur" – The doctor diagnosed me, "shukhistu 'ind ad-duktur bi-" – I was diagnosed by the doctor with, "shakhasuni mu'akhkharan bi-" – They recently diagnosed me with, "ana fi'lan shakhasuni bi-" – I was indeed diagnosed with

\subsection{Condition Keywords}

\textbf{Anxiety Disorders}
"idtirab al-qalaq" – Anxiety disorder, "al-qalaq" – Anxiety, "qalaq" – Worry, "at-tawatur" – Tension, "tawatur" – Stress

\textbf{Obsessive-Compulsive and Related Disorders}
"al-waswas al-qahri" – Obsessive-compulsive disorder, "waswas qahri" – OCD, "waswas" – Obsession, Obsessive-Compulsive Disorder, OCD

\textbf{Suicidal Thoughts and Self-Harm}
"afkar intihariya" – Suicidal thoughts, "fikr intihāri" – Suicidal ideation, Suicidal thoughts, "mayl lil-intihar" – Suicidal tendency, "intihar" – Suicide

\subsection{Code Snippets}
\label{snippets}
This is how we apply the keywords pattern given previously to identify diagnosed users.

\begin{lstlisting}[style=mypython, xleftmargin=0pt, xrightmargin=0pt]
#Looping through the diagnosis Keywords
#and Patterns


#Find occurrences of the diagnosis 
#phrase
phrase_matches = [(m.start(), m.end()) for m in re.finditer(re.escape(phrase), content)]


# Find all occurrences of the keyword
keyword_matches = [(m.start(), m.end()) for m in re.finditer(re.escape(keyword), content)]


# Check proximity between phrase and 
#keyword
for phrase_start, phrase_end in phrase_matches:
    for keyword_start, keyword_end in keyword_matches:
        if abs(phrase_start-keyword_start) <= char_range or abs(keyword_end-phrase_end) <= char_range:
            # Match Found
\end{lstlisting}

\subsection{Training Configuration}
\label{training-params}

We utilized publicly available pretrained models from Hugging Face for both fine-tuning and embeddings generation. The following models were employed:
\begin{itemize}
    \item \texttt{answerdotai/ModernBERT-base} \cite{modernbert}
    \item \texttt{aubmindlab/bert-base-arabertv02-twitter} (Twitter-optimized Arabic BERT) \cite{antoun2020arabert}
    \item \texttt{UBC-NLP/MARBERTv2} (updated Arabic social media model) \cite{abdul2021arbert}
    \item \texttt{CAMeL-Lab/bert-base-arabic-camelbert-mix} (state-of-the-art Arabic BERT) \cite{mubarak2023interplay}
    \item \texttt{microsoft/Multilingual-MiniLM-L12-H384} \cite{wang}
    \item \texttt{facebook/muppet-roberta-large} \cite{aghajanyan2021muppet}s
\end{itemize}

Following parameters were used to fine-tune and generate embeddings with publicly available PLMs from Hugging Face. Fine-tuning was executed on a series of \texttt{A100} GPUs.

\begin{lstlisting}[style=mypython, xleftmargin=0pt, xrightmargin=0pt]
eval_strategy="steps",
save_strategy="steps",
eval_steps=500,
save_steps=500,
learning_rate=2e-5,
per_device_train_batch_size=32,
per_device_eval_batch_size=32,
num_train_epochs=3,
weight_decay=0.01,
logging_steps=500,
load_best_model_at_end=True,
metric_for_best_model='weighted_f1',
fp16=True,
\end{lstlisting}

For predictive models in the hybrid pipeline, we used following configuration when making predictions based on the PLM embeddings:

\begin{lstlisting}[style=mypython, xleftmargin=0pt, xrightmargin=0pt]
"Logistic Regression": {
    "max_iter": 1000,
    "n_jobs": -1,
    "class_weight": "balanced"
},
"Random Forest": {
    "n_estimators": 200,
    "n_jobs": -1,
    "class_weight": "balanced_subsample"
},
"SVM": {
    "kernel": "linear",
    "probability": True,
    "class_weight": "balanced"
},
"MLP": {
    "hidden_layer_sizes": (512, 256),
    "max_iter": 500,
    "early_stopping": True,
    "alpha": 0.001
},
"XGBoost": {
    "n_estimators": 200,
    "max_depth": 6,
    "learning_rate": 0.1,
    "subsample": 0.8,
    "colsample_bytree": 0.8,
    "scale_pos_weight": "auto",
    "eval_metric": "logloss",
    "use_label_encoder": False,
    "tree_method": "gpu_hist" if torch.cuda.is_available() else "auto"
},
\end{lstlisting}


\end{document}